\documentclass[letterpaper, 10 pt, conference]{ieeeconf}  

\IEEEoverridecommandlockouts                              

\usepackage{amsmath,bm}
\usepackage{booktabs}
\usepackage{graphicx}
\usepackage{caption}
\usepackage{subcaption}
\usepackage{gensymb}
\usepackage{hyperref}
\hypersetup{
    colorlinks=true,
    linkcolor=red,
    filecolor=magenta,      
    urlcolor=magenta,
}
\usepackage{cite}

\newcommand{\vect}[1]{\mathbf{#1}}
\newcommand{\Lagr}{\mathcal{L}}
\newcommand{\mypara}[1]{\par\vspace*{1.5mm}\noindent\textbf{{#1}}}

\title{\LARGE \bf
Smile Like You Mean It: Driving Animatronic Robotic Face with Learned Models
}

\author{Boyuan Chen* \quad Yuhang Hu \quad Lianfeng Li \quad Sara Cummings \quad Hod Lipson\\
Columbia University
\thanks{We thank all the reviewers for their great comments and efforts. This research is supported by NSF NRI 1925157 and DARPA MTO grant HR0011-18-2-0020. *bchen@cs.columbia.edu}
}
\DeclareUnicodeCharacter{2212}{-}

\begin{document}

\maketitle
\thispagestyle{empty}
\pagestyle{empty}

\begin{abstract}

Ability to generate intelligent and generalizable facial expressions is essential for building human-like social robots. At present, progress in this field is hindered by the fact that each facial expression needs to be programmed by humans. In order to adapt robot behavior in real time to different situations that arise when interacting with human subjects, robots need to be able to train themselves without requiring human labels, as well as make fast action decisions and generalize the acquired knowledge to diverse and new contexts. We addressed this challenge by designing a physical animatronic robotic face with soft skin and by developing a vision-based self-supervised learning framework for facial mimicry. Our algorithm does not require any knowledge of the robot's kinematic model, camera calibration  or predefined expression set. By decomposing the learning process into a generative model and an inverse model, our framework can be trained using a single motor babbling dataset. Comprehensive evaluations show that our method enables accurate and diverse face mimicry across diverse human subjects.

\end{abstract}

\section{INTRODUCTION}

Facial expressions are an essential aspect of nonverbal communication. In our day-to-day lives, we rely on diverse facial expressions to convey our feelings and attitudes to others and interpret other people's emotions, desires and intentions \cite{plutchik1984emotions}. Facial mimicry \cite{reissland1988neonatal, meltzoff1989imitation, van2009love, piaget2013play} is also recognized as a vital stepping stone towards the early development of social skills for infants. Therefore, building robots that can automatically mimic diverse human facial expressions \cite{fong2003survey, blow2006art, breazeal2008social, saunderson2019robots} will facilitate more natural robotic social behaviors and further encourage stronger engagement in human-robot interactions. Mimicking human facial expressions is also the first step towards achieving adaptive facial reactions in robots. Despite the practical value of such systems, extant research in this domain mostly focuses on the hardware design and pre-programmed facial expressions, allowing robots to select one of the facial expressions from a predefined set. Generalizing across various human expressions has remained challenging.

\begin{figure}[t]
\begin{center}
    \includegraphics[width=.48\textwidth]{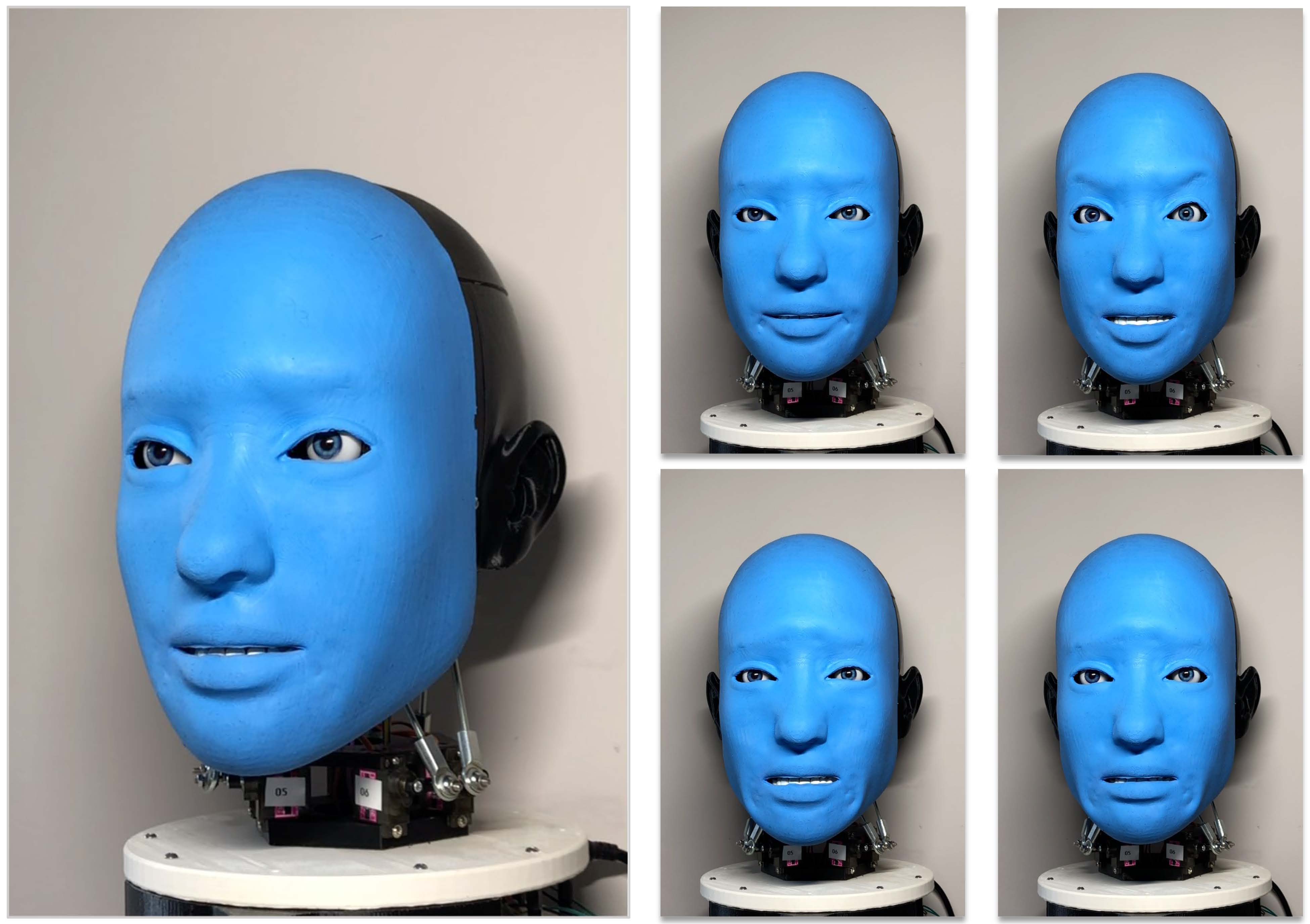}
\end{center}
\vspace{-3mm}
\caption{\textbf{Eva 2.0} is a general animatronic robotic face for facial mimicry. The robot does so by learning the correspondence between facial landmarks and self-images as well as a learned inverse kinematic model. The entire learning process relies on the robot's motor babbling in a self-supervised manner. Our robot can mimic varieties of human expressions across many human subjects.}
\label{fig:teaser}
\vspace{-13pt}
\end{figure}

Current robotic face systems cannot mimic human facial expressions adaptively. The key limitation is the lack of a general learning framework that can learn from limited human supervision. Some traditional methods \cite{oh2006design, hashimoto2006development,hashimoto2008dynamic, ahn2012designing, loza2013application, lin2016expressional, asheber2016humanoid} define a set of pre-specified facial expressions. Others generalize this process to search for closest match from a database \cite{liu2019emotion} or by following an fitness function \cite{hyung2019optimizing, liu2019emotion}. However, as human expressions are highly diverse, these approaches have limited value in practical robot−human interactions.

In this work, we present Eva 2.0 (Fig. \ref{fig:teaser}) with significant upgrades to our previous Eva 1.0 \cite{faraj2021facially} platform with more flexible and stable control. We further propose a general learning-based framework to learn facial mimicry from visual observations that can generalize well to different human subjects and diverse expressions. Importantly, our approach does not rely on human supervisions to provide ground-truth robot commands. Our key idea is to decompose the problem into two stages: (1) given normalized human facial landmarks, we first use a generative model to synthesize a corresponding robot self-image with the same facial expression and then (2) leverage an inverse network to output the set of motor commands from the synthesized image.

Our experiments suggest that our approach outperforms previous nearest-neighbor search-based algorithms and direct mapping from human face to action methods. Moreover, quantitative visualizations of our robot imitations demonstrate that, when presented with diverse human subjects, our method generates appropriate and accurate facial expression imitations.

Our primary contributions are threefold. First, we present an animatronic robotic face with soft skin and flexible control mechanisms. Second, we propose a vision-based learning framework for robot facial mimicry that can be trained in a self-supervised manner. Third, we construct a human facial expression dataset from previous database, YouTube videos and real-world human subjects. Our approach enables strong generalization over $12$ human subjects and nearly $400$ salient natural expressions. Our robot has a response time within 0.18 s, indicating a strong capability of real-time reaction. Qualitative results and open-sourced platforms are in our supplementary video at \url{https://rb.gy/9koqpg}.

\section{RELATED WORKS}

\mypara{Animatronic Robotics Face} Designing physical animatronic robot face capable of imitating human facial expressions is an important topic in human-robot interaction \cite{breazeal2003emotion, goodrich2008human, yan2014survey, kalegina2018characterizing, dimitrievska2020behavior, song2009image, liu2017facial, gu2017local}. Early works \cite{oh2006design, hashimoto2006development, itoh2006mechanical, hashimoto2008dynamic, asheber2016humanoid, loza2013application, ahn2012designing, lin2016expressional} solely focus on hardware design of the robot face and pre-program the facial expressions. Kismet \cite{brooks1998cog, breazeal1998toward, scassellati1998imitation, breazeal2000infant, breazeal2002regulation, breazeal2003emotion, breazeal2004designing} generates diverse facial expressions by interpolating among predefined basis facial postures over a three-dimensional space. Albert HUBO \cite{oh2006design} imitates specific human body movements, including facial expression. WE-4R \cite{itoh2006mechanical} determines the expression with emotion vectors. Hashimoto et al. \cite{hashimoto2008dynamic} maps the human face configuration to a robot through life-mask worn by a human subject. Asheber et al. \cite{asheber2016humanoid} proposed a simplified design to enable easier programming and flexible movements. However, all these designs rely on a fixed set of pre-programmed facial expressions that cannot generalize to novel expressions and require extensive human efforts through trail-and-error.

Recent studies start to integrate more general motion control into the hardware designs. Affetto \cite{ishihara2011realistic, ishihara2015design, ishihara2018identification} adopts hysterical sigmoid functions to model motor displacement. However, the landmarks are tracked with OptiTrack sensors on the face. Our method does not require any hardware attachment on the robot face. XIN-REN \cite{ren2016automatic} uses an AAM model to track the facial features for a specific person, which consequently cannot generalize well to novel subjects. They also calculate the robot solution with the robot kinematic equation, while ours learns the model automatically.

Hyung et al. \cite{hyung2019optimizing} have proposed a genetic algorithms to search for the best facial expression. However, the algorithm is inefficient when the search space grows with the complexity of the robot. Therefore, the approach cannot perform online inference, while our robot can react to a novel facial expression within 0.18s in a fully online manner. Liu and Ren \cite{liu2019emotion} classified the input emotion and search in a discrete pre-defined expression set which constrains possible outcomes.

\mypara{Synthetic Video Generation and Animation}
Significant progresses have been made in synthetic video generation with a focus on motion re-targeting. The goal of video synthesis \cite{wang2018video, wang2019few, siarohin2019animating, shaham2019singan} is to learn the mapping from a source video to a target video domain to render photorealisitc videos. Our work does not aim to render a realistic video or image. Rather, we aim to map human facial expressions to a physical robot platform.

On the other hand, some works in character face animation \cite{cao2014displaced, bulthoff2016perceptual, seymour2017interactive, nagano2018pagan, wei2019vr} targets at driving a digital face with human motions. Cao et al. \cite{cao2014displaced} learns to recognize human facial landmarks and map them to a 3D avatar. Their method requires full 3D knowledge of the digital face, making it hard to generalize to real robots. Our approach learns a kinematic model of the physical robot through self-supervised motor babbling without any prior knowledge of the control mechanisms.

The most relevant works pertain to motion re-targeting \cite{thies2015real, zhou2019dance, gafni2019vid2game, siarohin2019first}. Chan et al. \cite{chan2019everybody} proposes to transfer dance motion to a different human subject in a rendered video. Similarly, X2Face \cite{wiles2018x2face} and Face2Face \cite{thies2016face2face} learns to synthesize novel face movements videos. However, the aim of these works is realistic video rendering, rather than a physical robot realization. Moreover, most prior works assume a strong correlation of kinematic configurations between the source and target domain. However, as most robotic platforms, our robot does not share same control mechanisms as humans. Therefore, existing approaches cannot be directly applied in real-life contexts.

\mypara{Imitation Learning} Our work also shares the same high-level goal as imitation learning. However, most imitation learning research focus on the manipulation \cite{asfour2008imitation, rozo2013robot, zhang2018deep, ratliff2007imitation, muhlig2012interactive}, locomotion \cite{nakanishi2004learning, ratliff2007imitation, RoboImitationPeng20} or navigation \cite{8460487, driving-imitation, agile-off}, as discussed in detail in comprehensive review articles \cite{hussein2017imitation, osa2018algorithmic, 10.1145/3054912}. Our work studies facial mimicry. With our learning framework, we can directly train our models with the robotic face on an offline dataset and generalize on different human face configurations.

\section{DESIGN}

\begin{figure}[t]
\vspace*{5pt}
\begin{center}
    \includegraphics[width=.48\textwidth]{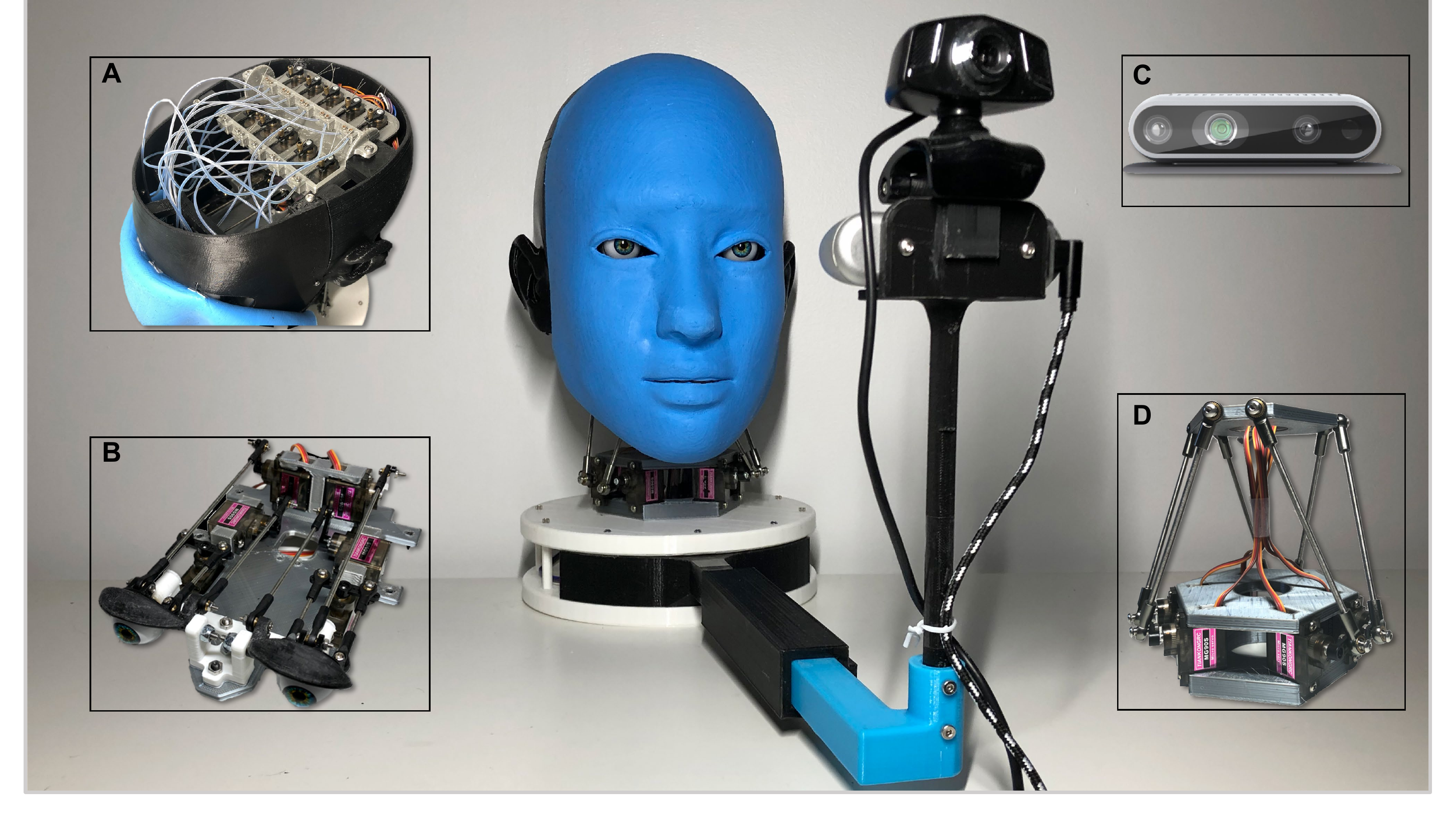}
\end{center}
\caption{\textbf{Mechanical Design:} our robot is actuated by the motor servo module (A) controlled by a Raspberry Pi 4 located at the bottom. The soft skin is connected to $10$ motors via nylon cord. Our 6DoF eye module (B) is decoupled from the front skull. The RGB camera (C) is only used for random data collection of robot self-images but not for testing. The 6DoF neck module (D) follows Steward platform.}
\label{fig:hardware}
\vspace{-13pt}
\end{figure}

Our robot design --- based on our previous Eva 1.0 \cite{faraj2021facially} with significant upgrades --- consists of two sub-assembly modules: facial movement module and neck movement module. We use micro servo motors (MG90S) to actuate all the components of our robotic face. All parts in our design are based on off-the-shelf hardware components that can be easily purchased online or 3D printed. To accelerate research, we open-source the design and step-by-step assembly process on our website. An overview of our hardware design is shown in Fig. \ref{fig:hardware}.

\mypara{Facial Movement Module} Our facial movement module can be further divided into skull frame, eye module, muscle module and jaw module. Compared to Eva 1.0, we redesign the 3D shape of the skull frame to enlarge the maneuvering space and thus allow for more flexible movements. The new skull frame also facilitates tighter connection to the skin with smoother and more natural looks on the face surface.

By adding a pair of ball joints and three pairs of parallelogram mechanisms, we now also ensure that the 6DoF eyeball module can move freely within a $\pm 20\degree$ range both horizontally and vertically, making the movements more stable. The rotation of each eyeball is controlled by two motors and three ball joint linkages. We also upgraded the eyelid design based on human face.

Our robot skin is attached to the front skull. The muscles driving facial expressions are attached to the inner side of the skin with a piece of fiber fabric, a strand of nylon cord, and a servo actuator inside the back of the skull. By pulling different nylon cords, a specific skin region can be deformed. We use the same region selection as Eva 1.0. To reduce the noise for muscle control, we run each nylon cord through a transparent vinyl tube linking the skull front and the motor. Our design ensures that the deformation in each muscle is proportional to the motor's rotation angle. Our design enables a large possible space of facial expressions by manipulating $10$ pairs of symmetrical muscles in different proportions.

The jaw module is responsible for deforming the two muscles around the mouth. This is similar to the aforementioned skin movement design, but benefits from two additional coupled motors that control the movement, allowing the jaw to open up to $20\degree$.

\mypara{Neck Movement Module} Our 6DoF neck module design is inspired by Stewart platform. Six motors are arranged in $3$ pairs evenly distributed on $3$ sides of the hexagon base. Each pair comprises two motors in a mirrored arrangement that are connected to a ball joint linkage.

\section{MODELS}

\begin{figure*}[!t]
    \vspace*{5pt}
    \centering
    \includegraphics[width=0.975\linewidth]{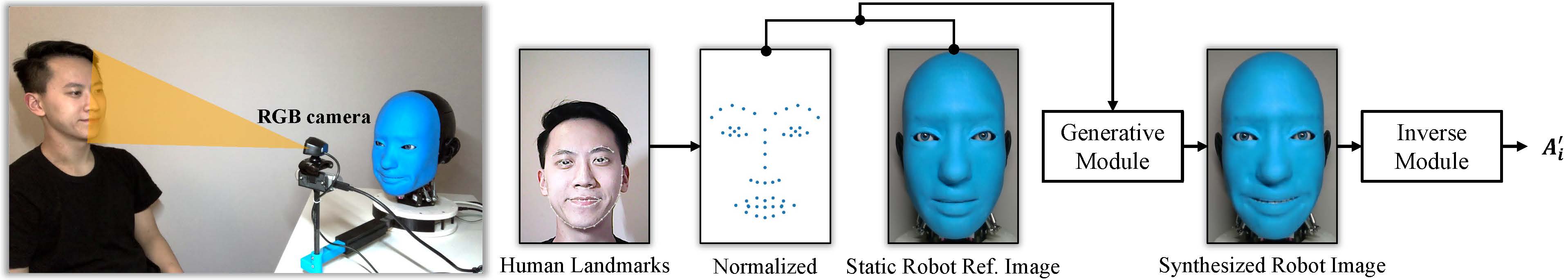}
    \caption{\textbf{Model Overview:} our two-stage framework consists of two major modules: a generative network and an inverse network. Given an image captured by a regular RGB camera, we first extract facial landmarks with OpenPose. We then normalize the human landmarks to the robot scale and embed it on an image. Together with a reference static robot self-image, these two images are concatenated to the generative network to synthesize a robot self-image as if the robot makes the same expression. The inverse model takes the synthetic robot self-image to output the final motor commands for execution.}
    \label{fig:overview}
    \vspace{-12pt}
\end{figure*}

We propose a learning-based framework for controlling the animatronic robotic face to mimic varieties of human facial expressions. An overview of our learning algorithms is shown in Fig. \ref{fig:overview}. We consider the following problem setup: given an image of human face displaying a natural facial expression, the model needs to output the motor commands to actuate the robot to imitate the given facial expression.

Without human pre-programming for different expressions, the problem poses several key challenges and desired properties. First, we hope that the learning algorithm can generalize to diverse unseen human faces. We leveraged the recent advances in human facial landmark detection to obtain abstract representations from high-dimensional image frames (Section \ref{representation}) which can be shared among different human subjects.

The second challenge in attempting to achieve facial mimicry stems from the lack of ground-truth pairs of human expressions and robot motor commands. Without hard-coding and extensive trail-and-error, obtaining such a pair is not practical. In this paper, we overcome this issue by adopting a two stage learning-based method: (1) a generative model which first synthesizes a robot self-image from facial landmarks processed by a proposed normalization algorithm (Section \ref{generative-model} and Section \ref{normalization}), (2) and an inverse model that is trained to produce desired motor commands from the generated robot image (Section \ref{inverse-model}). As we will show, the ground-truth labels for both models can be acquired with one round of self-supervised data collection without any human input.

\subsection{Representation of Facial Expression} \label{representation}

We capture facial expressions via facial landmarks, as this has been shown as an effective means of representing the underlying emotions. More importantly, facial landmarks provide a unified abstraction from diverse high-dimensional human images under different lighting, background and poses.

Specifically, we extract facial landmarks with OpenPose \cite{8765346, simon2017hand} software. The output is a vector of $53 \times 3$ size representing the spatial position of $53$ landmarks on human faces, with the last dimension being confidence scores. We also extracted the head pose for direct neck movement control. As ground-truth pairs for generating motor commands directly from the extracted human landmarks are not available, we adopt a different strategy, as discussed below.

\subsection{Landmark Normalization} \label{normalization}

Before we send the landmarks from human faces for robot learning, we need to normalize the spatial locations of the landmark vector to the robot domain. This is necessary due to potential variations in the scale or landmark arrangements in the captured human faces. Formally we normalize each landmark coordinate from human space $L_H$ to robot space $L_R$ with:
\begin{equation*}
    L_{R} = \frac{(L_{H}-H_{min})(R_{max} - R_{min})}{H_{max} - H_{min}} + R_{min}
\end{equation*}
where $H_{min}, H_{max}, R_{min}, R_{max}$ represent the value ranges of the spatial location per corresponding landmark in the sampled human and robot image frames.

\subsection{Generative Model} \label{generative-model}

The generative model takes in the normalized human landmarks and generates a synthetic RGB image allowing the robot to conceive the same facial expression, which we denote as robot's self-image. We parametrize the generative model $G$ with a deep neural network with parameter $\theta$.

A key challenge here is to map the coordinate vector to a high-dimensional image. This could be accomplished by encoding the input vectors with a fully-connected network, which is thus used by the robot to learn the entire spatial mapping. However, this simple approach requires a network of strong capacity and optimization algorithm and does not perform well in practice \cite{liu2018intriguing}.

To this end, we propose to encode the spatial coordinates of the landmarks to a two-channel image mask $\vect{M_{i}^{w \times h \times 2}}$ \cite{chan2019everybody}. The first channel has the value of $1$ if there is a landmark at the particular location and $0$ otherwise. The second channel is a greyscale image indicating the confidence score returned by the landmark detection algorithm. This encoding matches the size of the robot image, which helps with ensuring correct spatial correspondence.

Furthermore, instead of relying on the network to directly regress the absolute value in the output image, it only needs to output the “change” in the image introduced by landmark displacements. In practice, we achieve this by conditioning the network with a static robot self-image $\vect{I_{s}^{w \times h \times 3}}$ whereby the two images are concatenated along the depth channel. Our generative model can be expressed as: $\vect{I_i} \gets G(\vect{M_i}, \vect{I_s})$.

\mypara{Implementation Details}
We use a fully convolutional encoder-decoder architecture \cite{long2015fully, chen2021visual} where the resolution of the decoder network is enhanced by several hierarchical feature refinement convolutional layers \cite{dosovitskiy2015flownet}. Since the network is fully convolutional, we can preserve all the spatial information with high-quality outputs. Our network is optimized with a simple pixel-wise mean-squared error loss using Adam \cite{kingma2014adam} optimizer and a learning rate of 0.001. We train the network for 200 epochs with batch size $196$ until convergence on a validation dataset. We include more details in the supplementary. The formal objective function that we minimize is:
\begin{equation*}
    \Lagr_{G} = \text{MSE}(G(\vect{M}, \vect{I_s}), \vect{I})
\end{equation*}

\subsection{Inverse Model} \label{inverse-model}

The inverse model $F$ maps the synthetic robot self-image to motor commands in order to learn an inverse mapping from the goal image to actions.

\mypara{Action representation}
Our robot utilizes $N$ motors to actuate the face muscles. Given the robot face image, it is straightforward to frame this problem as continuous value regression. That is, given an input image, the network needs to output $N$ values for $N$ motor encoders. However, this framing can quickly become impractical in complex real-world scenarios and may lead to over-engineering. In practice, we observe that the motors have certain angle threshold to produce salient and stable motion. As a result, we can achieve salient motions with a discrete parameterized angle encoder without loss of accuracy.

We normalized and discretized the motor values to the $[0, 1]$ range with $0.25$ step size, resulting in $5$ values per motor. This discretization converts the original problem to a multi-class classification problem. Given an input robot image, the inverse model will output $5 \times N$ numbers, where every five numbers represents the probability of choosing each motor angle for one actuator. Our inverse model can be expressed as: $\vect{A_i} \gets F(\vect{I_i})$.

\mypara{Implementation Details}
Our architecture has $6$ convolutional layers followed by several fully-connected layers to adapt the output dimension to be $5 \times N$. We train our network with a multi-class Cross-Entropy loss with Adam optimizer and a learning rate of 0.00005 for $58$ epochs. We have $14,000$ pairs for training and $1,000$ pairs for validation and testing respectively. The formal objective function is:
\vspace{3pt}
\begin{equation*}
    \Lagr_F = \Sigma_{n=0}^{N-1} \text{CE}(F_{n}(\vect{I}), \vect{A_{n}})
\end{equation*}

\begin{figure}[t]
\begin{center}
    \includegraphics[width=.48\textwidth]{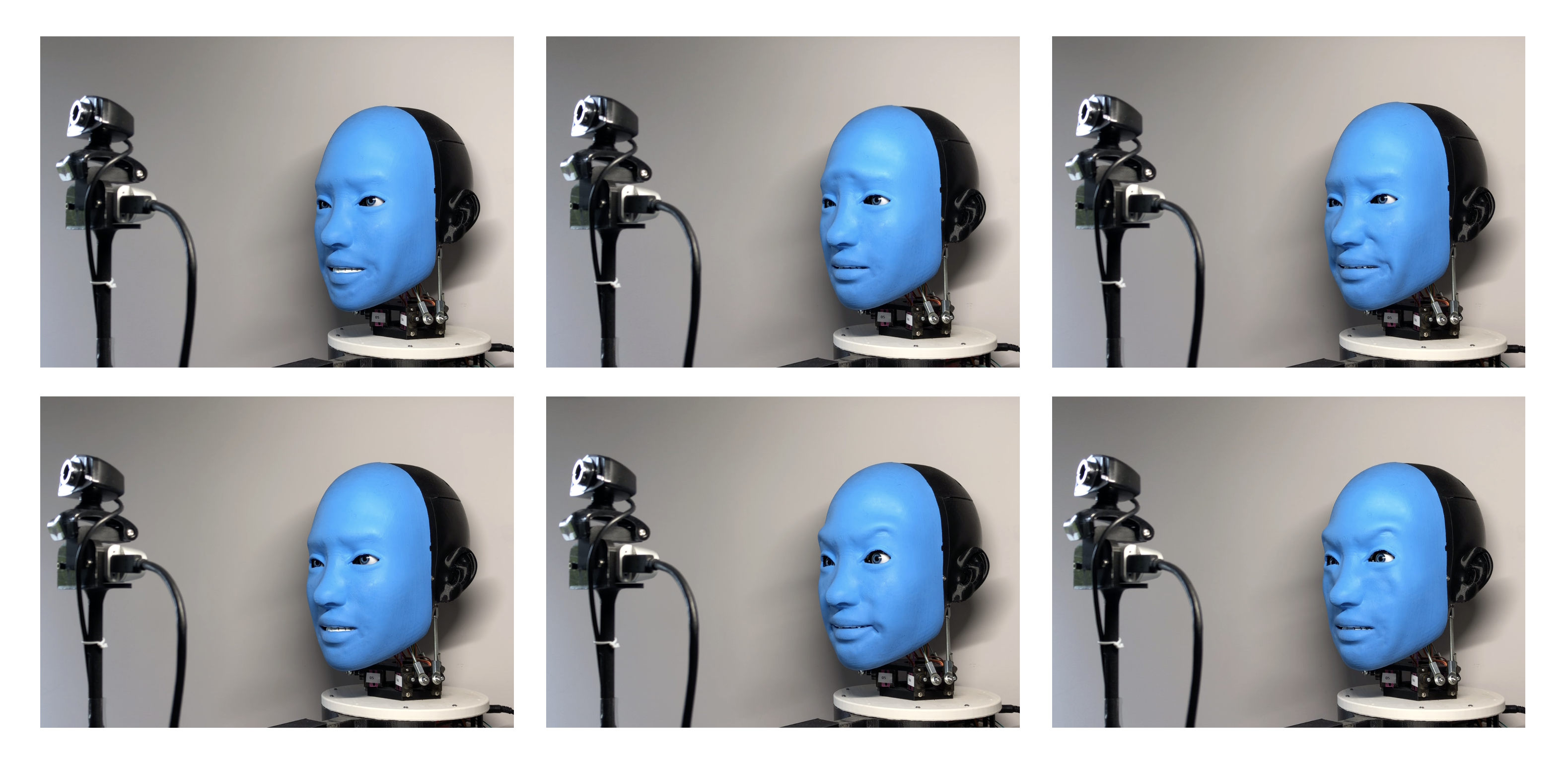}
\end{center}
\vspace{-3mm}
\caption{\textbf{Training Data Collection:} the whole training process of our generative model and the inverse model rely on a single robot dataset without human supervision. We collect our training data with random motor babbling in a self-supervised manner whereby the camera facing the robot is used solely for gathering the training data, \textit{i.e.}, it is not used during evaluation}
\vspace{-10pt}
\label{fig:robot-data-collection}
\end{figure}

\subsection{Training Data Collection} \label{training-data}

Both models can be trained separately with the same data collected via self-supervised motor babbling (Fig. \ref{fig:robot-data-collection}). We randomized the angles of the $10$ motors ranging from $0$ to $1$ with an interval of $0.25$ for $16,000$ steps. We used Intel RealSense D435i to capture RGB images and cropped the image to $480 \times 320$ to center the robot head. For each step $i$, we recorded the motor command values $\vect{A_i}$, the corresponding robot images $\vect{I_i}$, as well as the extracted landmark $\vect{L_{R, i}}$ from the robot with OpenPose. We thus obtain the training pairs of the generative model as ($\vect{L_{R, i}}$, $\vect{I_i}$) and the training pairs of the inverse model as ($\vect{I_i}$, $\vect{A_i}$). Since the data collection is purely random, the process does not require any human labeling.

\subsection{Inference}

Once the generative model and the inverse model are trained, we can use them jointly to perform inference. Given the input human image captured by an RGB camera, we first extract the landmark coordinates. After our normalization procedure, we can send it to our generative model which then outputs a synthetic robot self-image. This synthetic robot image serves as input for the inverse model which outputs the motor commands. All the network training and testing can be accomplished on a single NVIDIA 1080Ti GPU, whereas the motor commands on the robot are executed via WiFi.

\section{EXPERIMENTS}

\subsection{Evaluation Dataset}
Even though the training of our models do not require any human face dataset, we constructed a human facial dataset with salient facial expressions to perform extensive evaluations. Our human expression dataset is a combination of MMI Facial Expression database \cite{pantic2005web} and a set of online videos featuring eight subjects. We uniformly down-sampled the original videos to obtain $380$ salient frames covering a variety of facial expressions and human subjects.

\subsection{Baselines}

We are interested in evaluating the effectiveness of our approach for the generative model (GM), the inverse model (IM), and finally the entire pipeline. To this end, we design our baseline and ablation studies for each of them. For the generative model, we compared our method with a randomly sampled (RS) image from a collection of real images. This comparison aims to ascertain whether the synthetic image can outperform the real image and whether our model can predict reasonable motion on the static robot template.

Even though our method successfully convert the inverse model to a classification problem, a purely random baseline only has a $20\%$ success rate. We also compared our model with another two less random simple baselines --- a randomly initialized network (RI) and a model trained for about $100$ iterations (RI-100).

For the entire pipeline, we evaluated different combinations of the above baselines. For example, we obtained one baseline by combining our generative model with an inverse model that has been trained for $100$ iterations. Additionally, we present another three baselines. The first one is to perform a nearest neighbor retrieval (NN) with the landmarks extracted from the output of our generative model. We can directly search within our robot dataset. This baseline replaces our inverse model with a NN model. Similarly, we can perform such NN operation with human landmarks as a direct input. Lastly, we assessed whether we can directly generate the motor commands without our two-stage algorithm by training a network to output motor commands from our embedded input landmarks.

We used the same architecture for all the above baselines while varying only the number of channels in the first layer for different input dimensions. We trained all the models with three random seeds and report the mean and standard errors.

\subsection{Evaluation Metrics}

Our evaluation metrics cover both the pixel-wise accuracy of synthetic image and the accuracy of the output commands. We also extracted landmarks from our synthetic image to measure if the model successfully learns the facial expression than simply copying the static image. Moreover, we provide qualitative visualizations for the final pipeline. We also extracted landmarks from our entire pipeline execution to compare against the input human landmarks. We use L2 metric for both the image and landmark distances.

\subsection{Results}

\mypara{Generative Model}
Tab. \ref{tab:generative-only} shows the quantitative evaluation for our generative model. Our generative model outperforms the random baseline by a large margin. Note that the image distance is normalized by the total number of pixel values ($480 \times 320 \times 3$) which range from $0$ and $1$, whereas the landmark distance is normalized by the total number of landmarks ($53$).

\begin{table}[h!]
\caption{Accuracy of the Generative Model}
\label{tab:generative-only}
\centering
\resizebox{0.9\columnwidth}{!}{
\begin{tabular}{@{}l|c|c@{}}
\toprule
\multicolumn{1}{c|}{\textbf{Method}} & \textbf{\begin{tabular}[c]{@{}c@{}}Image Distance $\downarrow$ ($\times 10^{-5}$)\end{tabular}} & \multicolumn{1}{l}{\textbf{Landmark Distance $\downarrow$}} \\ \midrule
\textbf{GM (ours)}              & \bm{$3.47 \pm 0.009$}                                                            & \bm{$0.46 \pm 0.002$}                      \\ \midrule
RS                 & $6.47 \pm 0.039$                                                                     & $0.84 \pm 0.007$                               \\ \bottomrule
\end{tabular}
}
\vspace{-3pt}
\end{table}

\begin{figure*}[t!]
    \vspace*{5pt}
    \centering
    \includegraphics[width=0.975\linewidth]{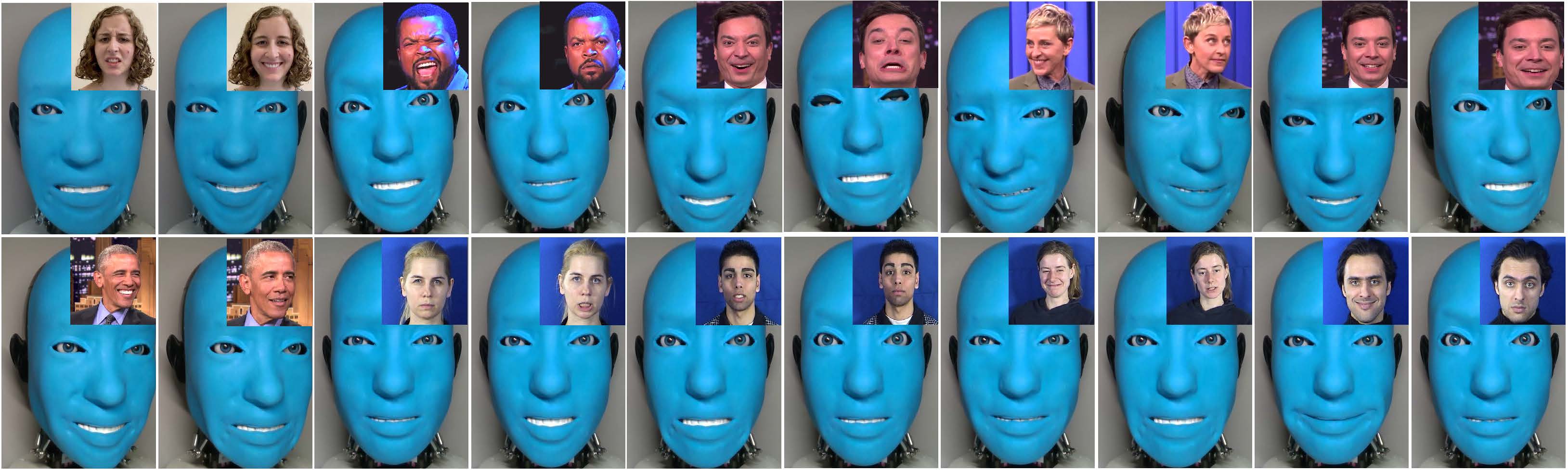}
    \caption{\textbf{Robot Visualizations:} we executed the output motor commands on our physical robot to demonstrate that our method supports accurate facial mimicry of a variety of human expressions across multiple human subjects.}
    \label{fig:visualizations}
    \vspace{-13pt}
\end{figure*}

\mypara{Inverse Model}
Tab. \ref{tab:inverse-only} shows the evaluation result for our inverse model. Compared with the random initialized model, the model trained for 100 iterations as well as the purely random baselines ($20\%$ accuracy), our inverse model produces much more accurate motor commands.

\begin{table}[h!]
\caption{Accuracy of the Inverse Model}
\label{tab:inverse-only}
\centering
\resizebox{0.9\columnwidth}{!}{
\begin{tabular}{@{}l|c|c@{}}
\toprule
\multicolumn{1}{c|}{\textbf{Method}} & \textbf{Command Distance $\downarrow$} & \multicolumn{1}{l}{\textbf{Command Accuracy $\uparrow$ (\%)}} \\ \midrule
\textbf{IM (ours)}                 & \bm{$0.53 \pm 0.009$} & \bm{$75.86 \pm 0.042$}                        \\ \midrule
RI                    & $1.39 \pm 0.009$          & $30.34 \pm 0.038$                                 \\ \midrule
RI-100         & $0.90 \pm 0.010$          & $56.40 \pm 0.041$                                 \\ \bottomrule
\end{tabular}
}
\vspace{-3pt}
\end{table}

\mypara{Two-Stage Pipeline}
By combining the generative and inverse model, we can use the synthetic output from the generative model as the input for the inverse model, allowing us to evaluate the entire two-stage pipeline. As shown in Tab. \ref{tab:two-stage}, our two-stage strategy outperforms all baselines, but is inferior to the performance of its component models. This is to be expected, as the inverse model takes the synthetic output from the generative model as its input when these are evaluated jointly, whereas individual evaluations are based on real images from our robot dataset. Nonetheless, our method still achieves the best predictions and is particularly advantageous compared to two single model baselines, indicating that decomposing the problem into two sub-steps is a prudent design choice.

\begin{table}[h!]
\caption{Accuracy of the Entire Two-Stage Pipeline}
\label{tab:two-stage}
\centering
\resizebox{\columnwidth}{!}{
\begin{tabular}{@{}cl|c|c@{}}
\toprule
\multicolumn{2}{c|}{\textbf{Method}}                                        & \textbf{\begin{tabular}[c]{@{}c@{}}Command Distance $\downarrow$ \end{tabular}}     & \textbf{\begin{tabular}[c]{@{}c@{}}Command Accuracy $\uparrow$ (\%)\end{tabular}} \\ \midrule
\multicolumn{1}{c|}{\textbf{GM}} & \multicolumn{1}{c|}{\textbf{IM}} &
\textbf{\begin{tabular}[c]{@{}c@{}}$\bm{0.83}$ $\bm{\pm 0.008}$\end{tabular}} & \textbf{\begin{tabular}[c]{@{}c@{}}$\bm{54.57}$ $\bm{\pm 0.048}$\end{tabular}} \\ \midrule
\multicolumn{1}{l|}{GM} & \multicolumn{1}{c|}{NN}                   & \begin{tabular}[c]{@{}c@{}}$1.06$ $\pm 0.009$\end{tabular}          & \begin{tabular}[c]{@{}c@{}}$45.74$ $\pm 0.045$\end{tabular}          \\ \midrule
\multicolumn{1}{l|}{GM} & \multicolumn{1}{c|}{RI}                  & \begin{tabular}[c]{@{}c@{}}$1.39$ $\pm 0.009$\end{tabular}          & \begin{tabular}[c]{@{}c@{}}$30.34$ $\pm 0.039$\end{tabular}          \\ \midrule
\multicolumn{1}{l|}{GM} & \multicolumn{1}{c|}{RI-100}               & \begin{tabular}[c]{@{}c@{}}$0.98$ $\pm 0.009$\end{tabular}          & \begin{tabular}[c]{@{}c@{}}$53.48$ $\pm 0.041$\end{tabular}          \\ \midrule
\multicolumn{2}{c|}{Landmark to Motor}                                      & \begin{tabular}[c]{@{}c@{}}$1.03$ $\pm 0.011$\end{tabular}         & \begin{tabular}[c]{@{}c@{}}$52.81$ $\pm 0.048$\end{tabular}          \\ \midrule
\multicolumn{2}{c|}{Landmark NN}                              &
\begin{tabular}[c]{@{}c@{}}$0.98$ $\pm 0.095$\end{tabular}         &
\begin{tabular}[c]{@{}c@{}}$49.3$ $\pm 0.045$\end{tabular}          \\ \bottomrule
\end{tabular}
\vspace{-17pt}
}
\end{table}

\mypara{Pipeline Execution}
We executed the output motor commands from the above two-stage pipeline on our physical robot and computed the landmark distance extracted from the resulted robot face with the ground-truth normalized landmarks. To provide qualitative evaluations, we visualized the final robot face together with the input human face in Fig. \ref{fig:visualizations}.

\begin{figure}[h!]
\begin{center}
    \includegraphics[width=.48\textwidth]{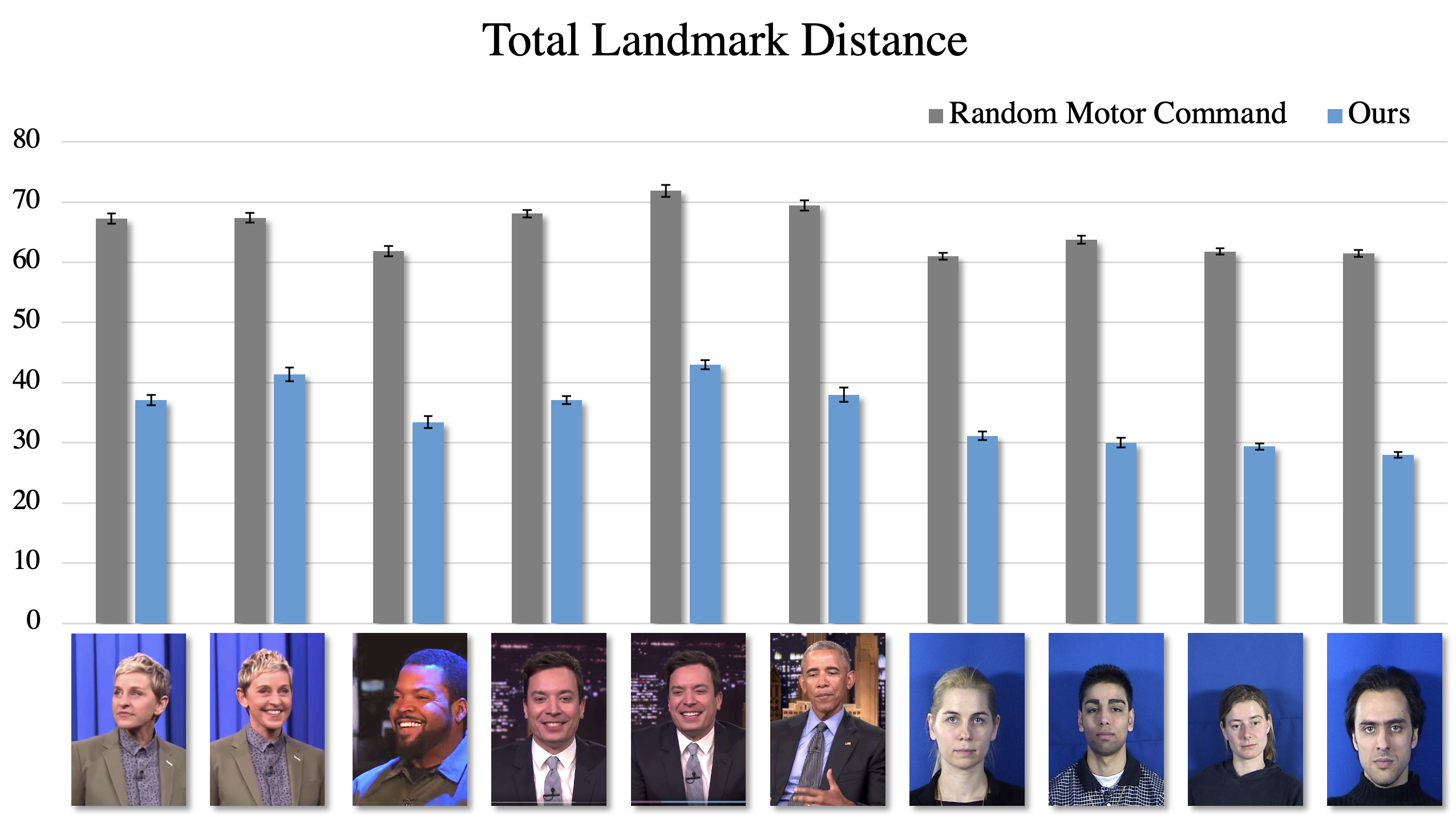}
\end{center}
\caption{\textbf{Pipeline Execution:} we executed the motor commands output by our entire two-stage pipeline and extracted the landmarks from the resulting physical robot face to compare against the ground-truth human landmarks as well as a random baseline to benchmark the task difficulty. The results show that our robot can imitate different human expressions accurately.}
\label{fig:10-bar-plot}
\vspace{-15pt}
\end{figure}

Our evaluation was conducted on $380$ salient frames which covers $10$ video clips from $8$ subjects. We show the quantitative results compared with a random baseline in Fig. \ref{fig:10-bar-plot}. Our approach demonstrates a flexible learning-based framework to mimic human facial expression. Our method also generalizes across various human subjects without any human supervision.

\section{CONCLUSIONS AND FUTURE WORK}

We present a new animatronic robot face design with soft skin and visual perception system. We also introduce a two-stage self-supervised learning framework for general face mimicry. Our experiments demonstrate that the two-stage algorithm improves the accuracy and diversity of imitating human facial expressions under various conditions. Our method enables real-time planning and opens new opportunities for practical applications.

Although we show that robots based on our design are capable of mimicking diverse human facial expressions from visual observations, it is necessary to study this problem with other modalities. For example, when several humans are involved in a conversation, the context will also influence their facial expressions. Thus, for future investigations, it would be beneficial to incorporate speech signal into the decision-making process while also allowing robots to reciprocate verbally.

Furthermore, even though imitation is an important step towards imbuing robots with more complex skills, being able to generate appropriate reaction expressions will be essential for interactive social robots. More broadly, building interactive robot face will require higher-level understanding of other's emotion, desires and intentions. Hence, an interesting direction is to explore higher-order thinking for robots.

\bibliographystyle{IEEEtran}
\bibliography{IEEEabrv, mybibfile}

\begin{thebibliography}{10}
\providecommand{\url}[1]{#1}
\csname url@rmstyle\endcsname
\providecommand{\newblock}{\relax}
\providecommand{\bibinfo}[2]{#2}
\providecommand\BIBentrySTDinterwordspacing{\spaceskip=0pt\relax}
\providecommand\BIBentryALTinterwordstretchfactor{4}
\providecommand\BIBentryALTinterwordspacing{\spaceskip=\fontdimen2\font plus
\BIBentryALTinterwordstretchfactor\fontdimen3\font minus
  \fontdimen4\font\relax}
\providecommand\BIBforeignlanguage[2]{{%
\expandafter\ifx\csname l@#1\endcsname\relax
\typeout{** WARNING: IEEEtran.bst: No hyphenation pattern has been}%
\typeout{** loaded for the language `#1'. Using the pattern for}%
\typeout{** the default language instead.}%
\else
\language=\csname l@#1\endcsname
\fi
#2}}

\bibitem{plutchik1984emotions}
R.~Plutchik, ``Emotions: A general psychoevolutionary theory,''
  \emph{Approaches to emotion}, vol. 1984, pp. 197--219, 1984.

\bibitem{reissland1988neonatal}
N.~Reissland, ``Neonatal imitation in the first hour of life: Observations in
  rural nepal.'' \emph{Developmental Psychology}, vol.~24, no.~4, p. 464, 1988.

\bibitem{meltzoff1989imitation}
A.~N. Meltzoff and M.~K. Moore, ``Imitation in newborn infants: Exploring the
  range of gestures imitated and the underlying mechanisms.''
  \emph{Developmental psychology}, vol.~25, no.~6, p. 954, 1989.

\bibitem{van2009love}
R.~Van~Baaren, L.~Janssen, T.~L. Chartrand, and A.~Dijksterhuis, ``Where is the
  love? the social aspects of mimicry,'' \emph{Philosophical Transactions of
  the Royal Society B: Biological Sciences}, vol. 364, no. 1528, pp.
  2381--2389, 2009.

\bibitem{piaget2013play}
J.~Piaget, \emph{Play, dreams and imitation in childhood}.\hskip 1em plus 0.5em
  minus 0.4em\relax Routledge, 2013, vol.~25.

\bibitem{fong2003survey}
T.~Fong, I.~Nourbakhsh, and K.~Dautenhahn, ``A survey of socially interactive
  robots,'' \emph{Robotics and autonomous systems}, vol.~42, no. 3-4, pp.
  143--166, 2003.

\bibitem{blow2006art}
M.~Blow, K.~Dautenhahn, A.~Appleby, C.~L. Nehaniv, and D.~Lee, ``The art of
  designing robot faces: Dimensions for human-robot interaction,'' in
  \emph{Proceedings of the 1st ACM SIGCHI/SIGART conference on Human-robot
  interaction}, 2006, pp. 331--332.

\bibitem{breazeal2008social}
C.~Breazeal, A.~Takanishi, and T.~Kobayashi, ``Social robots that interact with
  people.'' 2008.

\bibitem{saunderson2019robots}
S.~Saunderson and G.~Nejat, ``How robots influence humans: A survey of
  nonverbal communication in social human--robot interaction,''
  \emph{International Journal of Social Robotics}, vol.~11, no.~4, pp.
  575--608, 2019.

\bibitem{oh2006design}
J.-H. Oh, D.~Hanson, W.-S. Kim, Y.~Han, J.-Y. Kim, and I.-W. Park, ``Design of
  android type humanoid robot albert hubo,'' in \emph{2006 IEEE/RSJ
  International Conference on Intelligent Robots and Systems}.\hskip 1em plus
  0.5em minus 0.4em\relax IEEE, 2006, pp. 1428--1433.

\bibitem{hashimoto2006development}
T.~Hashimoto, S.~Hitramatsu, T.~Tsuji, and H.~Kobayashi, ``Development of the
  face robot saya for rich facial expressions,'' in \emph{2006 SICE-ICASE
  International Joint Conference}.\hskip 1em plus 0.5em minus 0.4em\relax IEEE,
  2006, pp. 5423--5428.

\bibitem{hashimoto2008dynamic}
T.~Hashimoto, S.~Hiramatsu, and H.~Kobayashi, ``Dynamic display of facial
  expressions on the face robot made by using a life mask,'' in \emph{Humanoids
  2008-8th IEEE-RAS International Conference on Humanoid Robots}.\hskip 1em
  plus 0.5em minus 0.4em\relax IEEE, 2008, pp. 521--526.

\bibitem{ahn2012designing}
H.~S. Ahn, D.-W. Lee, D.~Choi, D.-Y. Lee, M.~Hur, and H.~Lee, ``Designing of
  android head system by applying facial muscle mechanism of humans,'' in
  \emph{2012 12th IEEE-RAS International Conference on Humanoid Robots
  (Humanoids 2012)}.\hskip 1em plus 0.5em minus 0.4em\relax IEEE, 2012, pp.
  799--804.

\bibitem{loza2013application}
D.~Loza, S.~Marcos~Pablos, E.~Zalama~Casanova, J.~G{\'o}mez
  Garc{\'\i}a-Bermejo, and J.~L. Gonz{\'a}lez, ``Application of the facs in the
  design and construction of a mechatronic head with realistic appearance,''
  2013.

\bibitem{lin2016expressional}
C.-Y. Lin, C.-C. Huang, and L.-C. Cheng, ``An expressional simplified mechanism
  in anthropomorphic face robot design,'' \emph{Robotica}, vol.~34, no.~3, p.
  652, 2016.

\bibitem{asheber2016humanoid}
W.~T. Asheber, C.-Y. Lin, and S.~H. Yen, ``Humanoid head face mechanism with
  expandable facial expressions,'' \emph{International Journal of Advanced
  Robotic Systems}, vol.~13, no.~1, p.~29, 2016.

\bibitem{liu2019emotion}
N.~Liu and F.~Ren, ``Emotion classification using a cnn\_lstm-based model for
  smooth emotional synchronization of the humanoid robot ren-xin,'' \emph{PloS
  one}, vol.~14, no.~5, p. e0215216, 2019.

\bibitem{hyung2019optimizing}
H.-J. Hyung, H.~U. Yoon, D.~Choi, D.-Y. Lee, and D.-W. Lee, ``Optimizing
  android facial expressions using genetic algorithms,'' \emph{Applied
  Sciences}, vol.~9, no.~16, p. 3379, 2019.

\bibitem{faraj2021facially}
Z.~Faraj, M.~Selamet, C.~Morales, P.~Torres, M.~Hossain, B.~Chen, and
  H.~Lipson, ``Facially expressive humanoid robotic face,'' \emph{HardwareX},
  vol.~9, p. e00117, 2021.

\bibitem{breazeal2003emotion}
C.~Breazeal, ``Emotion and sociable humanoid robots,'' \emph{International
  journal of human-computer studies}, vol.~59, no. 1-2, pp. 119--155, 2003.

\bibitem{goodrich2008human}
M.~A. Goodrich and A.~C. Schultz, \emph{Human-robot interaction: a
  survey}.\hskip 1em plus 0.5em minus 0.4em\relax Now Publishers Inc, 2008.

\bibitem{yan2014survey}
H.~Yan, M.~H. Ang, and A.~N. Poo, ``A survey on perception methods for
  human--robot interaction in social robots,'' \emph{International Journal of
  Social Robotics}, vol.~6, no.~1, pp. 85--119, 2014.

\bibitem{kalegina2018characterizing}
A.~Kalegina, G.~Schroeder, A.~Allchin, K.~Berlin, and M.~Cakmak,
  ``Characterizing the design space of rendered robot faces,'' in
  \emph{Proceedings of the 2018 ACM/IEEE International Conference on
  Human-Robot Interaction}, 2018, pp. 96--104.

\bibitem{dimitrievska2020behavior}
V.~Dimitrievska and N.~Ackovska, ``Behavior models of emotion-featured robots:
  A survey,'' \emph{Journal of Intelligent \& Robotic Systems}, pp. 1--23,
  2020.

\bibitem{song2009image}
M.~Song, D.~Tao, Z.~Liu, X.~Li, and M.~Zhou, ``Image ratio features for facial
  expression recognition application,'' \emph{IEEE Transactions on Systems,
  Man, and Cybernetics, Part B (Cybernetics)}, vol.~40, no.~3, pp. 779--788,
  2009.

\bibitem{liu2017facial}
Z.~Liu, M.~Wu, W.~Cao, L.~Chen, J.~Xu, R.~Zhang, M.~Zhou, and J.~Mao, ``A
  facial expression emotion recognition based human-robot interaction system,''
  2017.

\bibitem{gu2017local}
J.~Gu, H.~Hu, and H.~Li, ``Local robust sparse representation for face
  recognition with single sample per person,'' \emph{IEEE/CAA Journal of
  Automatica Sinica}, vol.~5, no.~2, pp. 547--554, 2017.

\bibitem{itoh2006mechanical}
K.~Itoh, H.~Miwa, M.~Zecca, H.~Takanobu, S.~Roccella, M.~C. Carrozza, P.~Dario,
  and A.~Takanishi, ``Mechanical design of emotion expression humanoid robot
  we-4rii,'' in \emph{Romansy 16}.\hskip 1em plus 0.5em minus 0.4em\relax
  Springer, 2006, pp. 255--262.

\bibitem{brooks1998cog}
R.~A. Brooks, C.~Breazeal, M.~Marjanovi{\'c}, B.~Scassellati, and M.~M.
  Williamson, ``The cog project: Building a humanoid robot,'' in
  \emph{International Workshop on Computation for Metaphors, Analogy, and
  Agents}.\hskip 1em plus 0.5em minus 0.4em\relax Springer, 1998, pp. 52--87.

\bibitem{breazeal1998toward}
C.~Breazeal and J.~Vel{\'a}squez, ``Toward teaching a robot ‘infant’using
  emotive communication acts,'' in \emph{Proceedings of the 1998 Simulated
  Adaptive Behavior Workshop on Socially Situated Intelligence}.\hskip 1em plus
  0.5em minus 0.4em\relax Citeseer, 1998, pp. 25--40.

\bibitem{scassellati1998imitation}
B.~Scassellati, ``Imitation and mechanisms of joint attention: A developmental
  structure for building social skills on a humanoid robot,'' in
  \emph{International Workshop on Computation for Metaphors, Analogy, and
  Agents}.\hskip 1em plus 0.5em minus 0.4em\relax Springer, 1998, pp. 176--195.

\bibitem{breazeal2000infant}
C.~Breazeal and B.~Scassellati, ``Infant-like social interactions between a
  robot and a human caregiver,'' \emph{Adaptive Behavior}, vol.~8, no.~1, pp.
  49--74, 2000.

\bibitem{breazeal2002regulation}
C.~Breazeal, ``Regulation and entrainment in human—robot interaction,''
  \emph{The International Journal of Robotics Research}, vol.~21, no. 10-11,
  pp. 883--902, 2002.

\bibitem{breazeal2004designing}
C.~L. Breazeal, \emph{Designing sociable robots}.\hskip 1em plus 0.5em minus
  0.4em\relax MIT press, 2004.

\bibitem{ishihara2011realistic}
H.~Ishihara, Y.~Yoshikawa, and M.~Asada, ``Realistic child robot “affetto”
  for understanding the caregiver-child attachment relationship that guides the
  child development,'' in \emph{2011 IEEE International Conference on
  Development and Learning (ICDL)}, vol.~2.\hskip 1em plus 0.5em minus
  0.4em\relax IEEE, 2011, pp. 1--5.

\bibitem{ishihara2015design}
H.~Ishihara and M.~Asada, ``Design of 22-dof pneumatically actuated upper body
  for child android ‘affetto’,'' \emph{Advanced Robotics}, vol.~29, no.~18,
  pp. 1151--1163, 2015.

\bibitem{ishihara2018identification}
H.~Ishihara, B.~Wu, and M.~Asada, ``Identification and evaluation of the face
  system of a child android robot affetto for surface motion design,''
  \emph{Frontiers in Robotics and AI}, vol.~5, p. 119, 2018.

\bibitem{ren2016automatic}
F.~Ren and Z.~Huang, ``Automatic facial expression learning method based on
  humanoid robot xin-ren,'' \emph{IEEE Transactions on Human-Machine Systems},
  vol.~46, no.~6, pp. 810--821, 2016.

\bibitem{wang2018video}
T.-C. Wang, M.-Y. Liu, J.-Y. Zhu, G.~Liu, A.~Tao, J.~Kautz, and B.~Catanzaro,
  ``Video-to-video synthesis,'' \emph{arXiv preprint arXiv:1808.06601}, 2018.

\bibitem{wang2019few}
T.-C. Wang, M.-Y. Liu, A.~Tao, G.~Liu, J.~Kautz, and B.~Catanzaro, ``Few-shot
  video-to-video synthesis,'' \emph{arXiv preprint arXiv:1910.12713}, 2019.

\bibitem{siarohin2019animating}
A.~Siarohin, S.~Lathuili{\`e}re, S.~Tulyakov, E.~Ricci, and N.~Sebe,
  ``Animating arbitrary objects via deep motion transfer,'' in
  \emph{Proceedings of the IEEE Conference on Computer Vision and Pattern
  Recognition}, 2019, pp. 2377--2386.

\bibitem{shaham2019singan}
T.~R. Shaham, T.~Dekel, and T.~Michaeli, ``Singan: Learning a generative model
  from a single natural image,'' in \emph{Proceedings of the IEEE International
  Conference on Computer Vision}, 2019, pp. 4570--4580.

\bibitem{cao2014displaced}
C.~Cao, Q.~Hou, and K.~Zhou, ``Displaced dynamic expression regression for
  real-time facial tracking and animation,'' \emph{ACM Transactions on graphics
  (TOG)}, vol.~33, no.~4, pp. 1--10, 2014.

\bibitem{bulthoff2016perceptual}
H.~B{\"u}lthoff, C.~Wallraven, and M.~A. Giese, ``Perceptual robotics,'' in
  \emph{Springer Handbook of Robotics}.\hskip 1em plus 0.5em minus 0.4em\relax
  Springer, 2016, pp. 2095--2114.

\bibitem{seymour2017interactive}
M.~Seymour, K.~Riemer, and J.~Kay, ``Interactive realistic digital
  avatars-revisiting the uncanny valley,'' 2017.

\bibitem{nagano2018pagan}
K.~Nagano, J.~Seo, J.~Xing, L.~Wei, Z.~Li, S.~Saito, A.~Agarwal, J.~Fursund,
  and H.~Li, ``pagan: real-time avatars using dynamic textures,'' \emph{ACM
  Transactions on Graphics (TOG)}, vol.~37, no.~6, pp. 1--12, 2018.

\bibitem{wei2019vr}
S.-E. Wei, J.~Saragih, T.~Simon, A.~W. Harley, S.~Lombardi, M.~Perdoch,
  A.~Hypes, D.~Wang, H.~Badino, and Y.~Sheikh, ``Vr facial animation via
  multiview image translation,'' \emph{ACM Transactions on Graphics (TOG)},
  vol.~38, no.~4, pp. 1--16, 2019.

\bibitem{thies2015real}
J.~Thies, M.~Zollh{\"o}fer, M.~Nie{\ss}ner, L.~Valgaerts, M.~Stamminger, and
  C.~Theobalt, ``Real-time expression transfer for facial reenactment.''
  \emph{ACM Trans. Graph.}, vol.~34, no.~6, pp. 183--1, 2015.

\bibitem{zhou2019dance}
Y.~Zhou, Z.~Wang, C.~Fang, T.~Bui, and T.~Berg, ``Dance dance generation:
  Motion transfer for internet videos,'' in \emph{Proceedings of the IEEE
  International Conference on Computer Vision Workshops}, 2019, pp. 0--0.

\bibitem{gafni2019vid2game}
O.~Gafni, L.~Wolf, and Y.~Taigman, ``Vid2game: Controllable characters
  extracted from real-world videos,'' \emph{arXiv preprint arXiv:1904.08379},
  2019.

\bibitem{siarohin2019first}
A.~Siarohin, S.~Lathuili{\`e}re, S.~Tulyakov, E.~Ricci, and N.~Sebe, ``First
  order motion model for image animation,'' in \emph{Advances in Neural
  Information Processing Systems}, 2019, pp. 7137--7147.

\bibitem{chan2019everybody}
C.~Chan, S.~Ginosar, T.~Zhou, and A.~A. Efros, ``Everybody dance now,'' in
  \emph{Proceedings of the IEEE International Conference on Computer Vision},
  2019, pp. 5933--5942.

\bibitem{wiles2018x2face}
O.~Wiles, A.~Sophia~Koepke, and A.~Zisserman, ``X2face: A network for
  controlling face generation using images, audio, and pose codes,'' in
  \emph{Proceedings of the European conference on computer vision (ECCV)},
  2018, pp. 670--686.

\bibitem{thies2016face2face}
J.~Thies, M.~Zollhofer, M.~Stamminger, C.~Theobalt, and M.~Nie{\ss}ner,
  ``Face2face: Real-time face capture and reenactment of rgb videos,'' in
  \emph{Proceedings of the IEEE conference on computer vision and pattern
  recognition}, 2016, pp. 2387--2395.

\bibitem{asfour2008imitation}
T.~Asfour, P.~Azad, F.~Gyarfas, and R.~Dillmann, ``Imitation learning of
  dual-arm manipulation tasks in humanoid robots,'' \emph{International Journal
  of Humanoid Robotics}, vol.~5, no.~02, pp. 183--202, 2008.

\bibitem{rozo2013robot}
L.~Rozo, P.~Jim{\'e}nez, and C.~Torras, ``A robot learning from demonstration
  framework to perform force-based manipulation tasks,'' \emph{Intelligent
  service robotics}, vol.~6, no.~1, pp. 33--51, 2013.

\bibitem{zhang2018deep}
T.~Zhang, Z.~McCarthy, O.~Jow, D.~Lee, X.~Chen, K.~Goldberg, and P.~Abbeel,
  ``Deep imitation learning for complex manipulation tasks from virtual reality
  teleoperation,'' in \emph{2018 IEEE International Conference on Robotics and
  Automation (ICRA)}.\hskip 1em plus 0.5em minus 0.4em\relax IEEE, 2018, pp.
  1--8.

\bibitem{ratliff2007imitation}
N.~Ratliff, J.~A. Bagnell, and S.~S. Srinivasa, ``Imitation learning for
  locomotion and manipulation,'' in \emph{2007 7th IEEE-RAS International
  Conference on Humanoid Robots}.\hskip 1em plus 0.5em minus 0.4em\relax IEEE,
  2007, pp. 392--397.

\bibitem{muhlig2012interactive}
M.~M{\"u}hlig, M.~Gienger, and J.~J. Steil, ``Interactive imitation learning of
  object movement skills,'' \emph{Autonomous Robots}, vol.~32, no.~2, pp.
  97--114, 2012.

\bibitem{nakanishi2004learning}
J.~Nakanishi, J.~Morimoto, G.~Endo, G.~Cheng, S.~Schaal, and M.~Kawato,
  ``Learning from demonstration and adaptation of biped locomotion,''
  \emph{Robotics and autonomous systems}, vol.~47, no. 2-3, pp. 79--91, 2004.

\bibitem{RoboImitationPeng20}
X.~B. Peng, E.~Coumans, T.~Zhang, T.-W.~E. Lee, J.~Tan, and S.~Levine,
  ``Learning agile robotic locomotion skills by imitating animals,'' in
  \emph{Robotics: Science and Systems}, 07 2020.

\bibitem{8460487}
F.~{Codevilla}, M.~{Müller}, A.~{López}, V.~{Koltun}, and A.~{Dosovitskiy},
  ``End-to-end driving via conditional imitation learning,'' in \emph{2018 IEEE
  International Conference on Robotics and Automation (ICRA)}, 2018, pp.
  4693--4700.

\bibitem{driving-imitation}
\BIBentryALTinterwordspacing
F.~Codevilla, M.~M{\"{u}}ller, A.~Dosovitskiy, A.~M. L{\'{o}}pez, and
  V.~Koltun, ``End-to-end driving via conditional imitation learning,''
  \emph{CoRR}, vol. abs/1710.02410, 2017. [Online]. Available:
  \url{http://arxiv.org/abs/1710.02410}
\BIBentrySTDinterwordspacing

\bibitem{agile-off}
\BIBentryALTinterwordspacing
Y.~Pan, C.~Cheng, K.~Saigol, K.~Lee, X.~Yan, E.~A. Theodorou, and B.~Boots,
  ``Agile off-road autonomous driving using end-to-end deep imitation
  learning,'' \emph{CoRR}, vol. abs/1709.07174, 2017. [Online]. Available:
  \url{http://arxiv.org/abs/1709.07174}
\BIBentrySTDinterwordspacing

\bibitem{hussein2017imitation}
A.~Hussein, M.~M. Gaber, E.~Elyan, and C.~Jayne, ``Imitation learning: A survey
  of learning methods,'' \emph{ACM Computing Surveys (CSUR)}, vol.~50, no.~2,
  pp. 1--35, 2017.

\bibitem{osa2018algorithmic}
T.~Osa, J.~Pajarinen, G.~Neumann, J.~A. Bagnell, P.~Abbeel, and J.~Peters, ``An
  algorithmic perspective on imitation learning,'' \emph{arXiv preprint
  arXiv:1811.06711}, 2018.

\bibitem{10.1145/3054912}
\BIBentryALTinterwordspacing
A.~Hussein, M.~M. Gaber, E.~Elyan, and C.~Jayne, ``Imitation learning: A survey
  of learning methods,'' \emph{ACM Comput. Surv.}, vol.~50, no.~2, Apr. 2017.
  [Online]. Available: \url{https://doi.org/10.1145/3054912}
\BIBentrySTDinterwordspacing

\bibitem{8765346}
Z.~{Cao}, G.~{Hidalgo Martinez}, T.~{Simon}, S.~{Wei}, and Y.~A. {Sheikh},
  ``Openpose: Realtime multi-person 2d pose estimation using part affinity
  fields,'' \emph{IEEE Transactions on Pattern Analysis and Machine
  Intelligence}, 2019.

\bibitem{simon2017hand}
T.~Simon, H.~Joo, I.~Matthews, and Y.~Sheikh, ``Hand keypoint detection in
  single images using multiview bootstrapping,'' in \emph{CVPR}, 2017.

\bibitem{liu2018intriguing}
R.~Liu, J.~Lehman, P.~Molino, F.~P. Such, E.~Frank, A.~Sergeev, and
  J.~Yosinski, ``An intriguing failing of convolutional neural networks and the
  coordconv solution,'' in \emph{Advances in Neural Information Processing
  Systems}, 2018, pp. 9605--9616.

\bibitem{long2015fully}
J.~Long, E.~Shelhamer, and T.~Darrell, ``Fully convolutional networks for
  semantic segmentation,'' in \emph{Proceedings of the IEEE conference on
  computer vision and pattern recognition}, 2015, pp. 3431--3440.

\bibitem{chen2021visual}
B.~Chen, C.~Vondrick, and H.~Lipson, ``Visual behavior modelling for robotic
  theory of mind,'' \emph{Scientific Reports}, vol.~11, no.~1, pp. 1--14, 2021.

\bibitem{dosovitskiy2015flownet}
A.~Dosovitskiy, P.~Fischer, E.~Ilg, P.~Hausser, C.~Hazirbas, V.~Golkov, P.~Van
  Der~Smagt, D.~Cremers, and T.~Brox, ``Flownet: Learning optical flow with
  convolutional networks,'' in \emph{Proceedings of the IEEE international
  conference on computer vision}, 2015, pp. 2758--2766.

\bibitem{kingma2014adam}
D.~P. Kingma and J.~Ba, ``Adam: A method for stochastic optimization,''
  \emph{arXiv preprint arXiv:1412.6980}, 2014.

\bibitem{pantic2005web}
M.~Pantic, M.~Valstar, R.~Rademaker, and L.~Maat, ``Web-based database for
  facial expression analysis,'' in \emph{2005 IEEE international conference on
  multimedia and Expo}.\hskip 1em plus 0.5em minus 0.4em\relax IEEE, 2005, pp.
  5--pp.

\end{thebibliography}

\end{document}